\documentclass[sigconf, nonacm, 10pt]{acmart}

\renewcommand\footnotetextcopyrightpermission[1]{} 

\usepackage{enumitem}
\usepackage{epsfig}
\usepackage{graphicx}
\usepackage{amsmath}
\usepackage{arydshln}
\usepackage{url}            
\usepackage{booktabs}       
\usepackage{amsfonts}       
\usepackage{nicefrac}       
\usepackage{microtype}      
\usepackage{bbding}
\usepackage[linesnumbered, ruled, vlined]{algorithm2e}
\usepackage{adjustbox}
\usepackage{multirow}
\usepackage{wrapfig}
\usepackage{fancyhdr}

\newcommand{\specialcellC}[2][c]{%
  \begin{tabular}[#1]{@{}c@{}}#2\end{tabular}}     

\AtBeginDocument{%
  \providecommand\BibTeX{{%
    \normalfont B\kern-0.5em{\scshape i\kern-0.25em b}\kern-0.8em\TeX}}}

\setcopyright{rightsretained}
\copyrightyear{2023}
\acmYear{2023}
\acmDOI{10.1145/3578356.3592579}

\begin{document}

\title{Gradient-less Federated Gradient Boosting Trees\\ with Learnable Learning Rates}



\author{Chenyang Ma, Xinchi Qiu, Daniel J. Beutel, Nicholas D. Lane}
\affiliation{%
  \institution{University of Cambridge}
  \city{}
  \country{}    
  }
\email{cm2196@cam.ac.uk}






\begin{abstract}
The privacy-sensitive nature of decentralized datasets and the robustness of eXtreme Gradient Boosting (XGBoost) on tabular data raise the needs to train XGBoost in the context of federated learning (FL). Existing works on federated XGBoost in the horizontal setting rely on the sharing of gradients, which induce per-node level communication frequency and serious privacy concerns. To alleviate these problems, we develop an innovative framework for horizontal federated XGBoost which does not depend on the sharing of gradients and simultaneously boosts privacy and communication efficiency by making the learning rates of the aggregated tree ensembles learnable. We conduct extensive evaluations on various classification and regression datasets, showing our approach achieves performance comparable to the state-of-the-art method and effectively improves communication efficiency by lowering both communication rounds and communication overhead by factors ranging from 25x to 700x.
\end{abstract}

\maketitle
\pagestyle{plain} 

\section{Introduction}

Federated Learning (FL) enables the training of a global model using decentralized datasets in a privacy-preserving manner, contrasting with the conventional centralized training paradigm \cite{yang2024unitouch, yang2022touch, yang2023generating, ma2024see, dou2024tactile}. Existing FL research \cite{McMahanMRHA17, LiSZSTS20, ReddiCZGRKKM21, LiJZKD21} and the developed techniques \cite{Kairouz21_Open_Problems, BonawitzEGHIIKK19, Wu_2023_boosting, Zhao2022RBCRT} to optimize model convergence and reduce systematic privacy risks and costs mainly focus on neural networks (NN). The efforts of developing FL algorithms to support other machine learning (ML) models, on the other hand, remain under-explored.

EXtreme Gradient Boosting (XGBoost) \cite{XGBoost} is a powerful and interpretable gradient-boosted decision tree (GBDT). In most of the cases, XGBoost outperforms deep learning methods for tabular data on medium-sized datasets under 10k training examples \cite{XGBoost_outperform_DL, SHWARTZZIV202284, yang2022sparse}. In the context of cross-silo FL, where the clients pool is typically made up from 2 to 100 \cite{Kairouz21_Open_Problems} organizations, there is a growing need to deploy a federated XGBoost system on specific tasks such as survival analysis \cite{survival_Barnwal} and financial fraud detection \cite{Secureboost_Cheng, WangQL0JWFYZY19}.

\begin{figure}[t]
\setlength{\abovecaptionskip}{0.1cm}
\begin{center}
\includegraphics[width=0.95\linewidth]{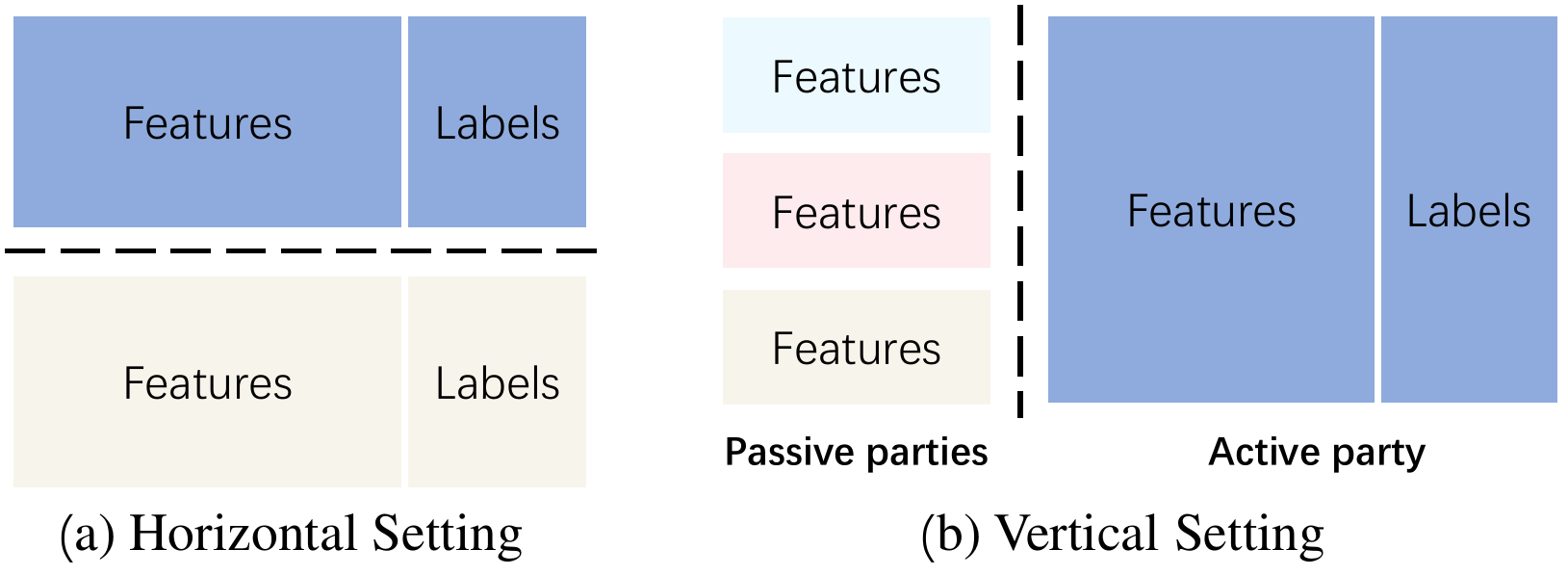}
\end{center}
  \caption{Horizontal vs. vertical federated XGBoost.}
  \label{Teaser}
\end{figure}

Existing works on federated XGBoost usually follow two settings (Fig. \ref{Teaser}). The horizontal setting is defined to be the case when clients' datasets have identical feature spaces but different sample IDs. The central server sends the global model to all clients and then aggregates the updated model parameters after each communication round. As for the vertical setting, first proposed by SecureBoost \cite{Secureboost_Cheng, QiuvFedSec}, the concepts of passive parties and one active party were introduced, where passive parties and the active party share identical sample space but possess different features. As only the active party owns data labels, it naturally acts as the server.

Although the horizontal setting remains to be more common \cite{Samplwise_Jones}, the training of a horizontal federated XGBoost turns out to be harder, not easier, because finding the optimal split condition of XGBoost trees depends on the \textit{order} of the data samples \cite{Tian2020FederBoostPF, QiuEVFLSA} as we iterate the feature set and partition the data samples into left and right according to the feature constraints. Therefore, as all clients share the same sample IDs in the vertical setting, the passive parties only need to send the order of the samples to the active party. However, since the sample IDs are different across all clients in the horizontal setting, at every splitting point, each client needs to transmit the gradients, hessians, and/or sample splits based on the feature values to the server to find the optimal splitting condition \cite{Adaptive_Ong}. Hence, we identify two key problems imposed by this vanilla approach.

\begin{itemize}[leftmargin=*]
\item[1.] \textbf{\textit{Per-node level communication frequency.}} The server needs to communicate with all clients at every splitting point. We denote the depth of each tree as $L$ and the number of trees in the tree ensemble as $M$. The number of nodes in the tree ensemble can scale up to $M \times 2^{L}$ \cite{Chang0Z22}, and so is the number of communication rounds. As a trained XGBoost model is common to have a depth of 8 and 500 trees \cite{XGBoost}, the number of communication rounds can reach $\sim$100K. Moreover, in a real application of federated XGBoost, it is possible for the server to conduct more than one round of communication per node \cite{LiuMLMND020} and carry out extra cryptographic calculations. Thus, the high communication overhead makes it difficult to deploy horizontal federated XGBoost for practical uses.

\item[2.] \textbf{\textit{Serious privacy concerns.}} The sharing of gradients and even confident information was proved to be insecure in the distributed training of ML models \cite{ZhuLH19, FredriksonJR15}. As the training data can be reconstructed using gradients, such sharing needs to be protected.
\end{itemize}

Existing research on horizontal federated XGBoost tackles the aforementioned two problems by seeking a trade-off between privacy and communication costs. A few works take stronger defenses against privacy leaks. FedXGB \cite{LiuMLMND020} developed a new secure aggregation protocol by applying homomorphic encryption and secret sharing on shared parameters directly. However, this induces high communication and computation overhead at per-node level communication frequency. Some works decrease the resolution of the raw data distribution by generating a surrogate representation using gradient histogram \cite{Tian2020FederBoostPF, Chang0Z22, Samplwise_Jones, Adaptive_Ong}. Histogram-based methods accelerate the training process by building quantile sketch approximation, but the communication frequency still correlates to the depth of the trees. Besides, they can still leak privacy because the gradients related to the bins and the thresholds can be inferred \cite{YamamotoOW22}. Other works obfuscate the raw data distribution with methods including clustering-based k-anonymity \cite{kanonymity_Yang} and locality-sensitive hashing \cite{practical_LiWH20}. Although the required communication overhead is less than encryption-based methods, these approaches have a trade-off between model performance and the number of clients.

In this work, we ask the fundamental question: \textbf{\textit{if it is possible not to rely on the sharing of gradients and hessians to construct a federated XGBoost in the horizontal setting?}} In this way, we can simultaneously boost privacy and disentangle the per-node level communication frequency. We find it to be possible by formulating an important intuition: as the local datasets of clients can be heterogeneous in the horizontal setting, using a fixed learning rate for each tree may be too weak since each tree can make different amounts of mistakes on unseen data with distribution shifts. To this end, we make the learning rates of the aggregated tree ensembles learnable by training a small one-layer 1D CNN with kernel size and stride equal to the number of trees in each client tree ensemble. We use the prediction outcomes as inputs directly. This novel framework preserves privacy. The clients only need to send the constructed tree ensemble to the server. The sharing of gradients and hessians, which may leak sensitive information, is not required. In addition, the number of communication rounds is independent of any hyperparameter related to the trained XGBoost. In practice, we find 10 communication rounds to be sufficient for the global federated XGBoost model to reach performance comparable to the state-of-the-art method. Moreover, the total communication overhead to train a global federated XGBoost model is independent of the dataset size. Our approach induces total communication overhead lower than previous works in the order of tens to hundreds. 

The main contributions of this work are summarized as:

\begin{itemize}[leftmargin=*]
\item We propose a novel privacy-preserving framework, \textbf{FedXGBllr}, a \textbf{fed}erated \textbf{XGB}oost with \textbf{l}earnable \textbf{l}earning \textbf{r}ates in the horizontal setting which do not rely on the sharing of gradients and hessians.
\item Our framework disentangles the per-node level communication frequency when training a federated XGBoost.
\item The total communication overhead of our framework is independent of the dataset size and is significantly lower (by factors ranging from  25x to 700x) than previous methods.
\item We show that FedXGBllr is interpretable with carefully framed reasoning and analysis.
\item We conduct extensive experiments on both classification and regression datasets with diverse feature dimensions and sizes. Experimental results of our approach are comparable to the state-of-the-art horizontal federated XGBoost.
\item The implementation is \href{https://flower.ai/blog/2023-04-19-xgboost-with-flower/}{\textcolor{magenta}{made available}} and released under Flower \cite{FlowerBeutel}, a comprehensive, end-to-end FL framework.
\end{itemize}

\section{Preliminaries}\label{Section2}
\textbf{EXtreme Gradient Boosting (XGBoost)} \hspace{1.0mm} XGBoost is a gradient boosting tree and is an additive ensemble model. It adopts forward stagewise regression and consistently learns new trees to fit the residuals until a stop condition is met. Given a dataset $\{x_i, y_i\}_{i=1}^N$ where $x_i\in\mathbb{R}^D, y_i\in\mathbb{R}$ represent the features (with dimension $D$) and labels of the $i$-th sample, the final prediction is calculated by summing predictions of all $M$ trees with a fixed learning rate $\eta$:
\begin{equation}\label{eq1}
\hat{y}_i = \sum_{t=1}^{M} \eta f_t(x_i)
\end{equation}

\noindent where $f_t(x_i)$ is the prediction made by the $t$-th tree. 

The objective of XGBoost is to minimize the sum of the loss of all the data samples. It first calculates the first order-gradient, $g_i$, and second-order hessian, $h_i$, of all samples:
\begin{equation}\label{eq2}
\begin{gathered}
g_i = {\partial_{\hat{y}_i^{(t-1)}}}L(y_i,\hat{y}_i^{t-1}), \;\;\;
h_i = {\partial_{\hat{y}_i^{(t-1)}}^{2}}L(y_i,\hat{y}_i^{t-1})
\end{gathered}
\end{equation}

\noindent where $\hat{y}_i^{(t-1)}$ is the prediction made by the previous tree and $L(y_i,\hat{y}_i^{t-1})$ is the loss function. Then the gradient sums of instance set $I_j$ on each node $j$ can be calculated by:
\begin{equation}\label{eq3}
G_j = \sum_{i \in I_j}g_i, \;\;\; H_j = \sum_{i \in I_j}h_i 
\end{equation}

The optimal weight $w_j^*$ and objective $obj^*$ are derived from the objective function involving regularization terms:
\begin{equation}\label{eq4}
\begin{aligned}
w_{j}^{*} = -\frac{G_j}{H_{j} + \lambda}, \;\;\;
obj^{*} = -\frac{1}{2} \sum_{j=1}^{T} \frac{G_j^2}{H_{j} + \lambda} + \gamma{T}
\end{aligned}
\end{equation}

\noindent where $T$ is the leaf node number and $\lambda$ and $\gamma$ are the regularization for the leaf weights and leaf number, respectively. 

From root to leaf nodes, the best split can be found by maximizing $SplitGain = obj_{before}^{*} - obj_{after}^{*}$, which is:
\begin{equation}\label{eq5}
SplitGain = \frac{1}{2}[ \ \frac{G_L^2}{H_{L} + \lambda} + \frac{G_R^2}{H_{R} + \lambda} - \frac{G_L^2 + G_R^2}{H_{L} + H_{R} + \lambda} ] \ - \gamma
\end{equation}

\noindent where $G_L$ and $H_L$, $G_R$, and $H_R$ are the sums of the gradients and hessians of the data samples partitioned into the left and right branch based on the splitting point's feature constraint.

\section{Method}
In this section, we provide a detailed description of our approach. We first formulate our intuitions in Section \ref{Section3.1}. We then facilitate our intuitions in Section \ref{Section3.2} and discuss how to learn the learning rates using proposed, interpretable one-layer 1D CNN in Section \ref{Section3.3}. Finally, we develop new framework \textbf{FedXGBllr} to train federated XGBoost in Section \ref{Section3.4}.

\subsection{Intuitions} \label{Section3.1}

\textbf{A fixed learning rate is too weak} \hspace{1.0mm} Local datasets of clients participating in FL can be heterogeneous (i.e., non-IID). The trained model on the client's local dataset converges to its local optima. When the model is sent to other clients and evaluated on their local datasets, it suffers from degradation in performance because different clients' local optima are divergent. The adverse effects of data heterogeneity in FL over NN-based approaches are widely researched \cite{LiHYWZ20, LiDCH22, KarimireddyKMRS20, LiSZSTS20}. More recent works demonstrate that XGBoost also experiences deterioration in model performance with heterogeneous local datasets \cite{Samplwise_Jones, FanXF020}.

We argue that the core reason causing performance degradation, when the built XGBoost model is evaluated on other unseen datasets with distribution shifts, is that each tree in the tree ensemble makes different amounts of mistakes. 

\begin{figure}[t]
\setlength{\abovecaptionskip}{0.1cm}
\begin{center}
\includegraphics[width=0.95\linewidth]{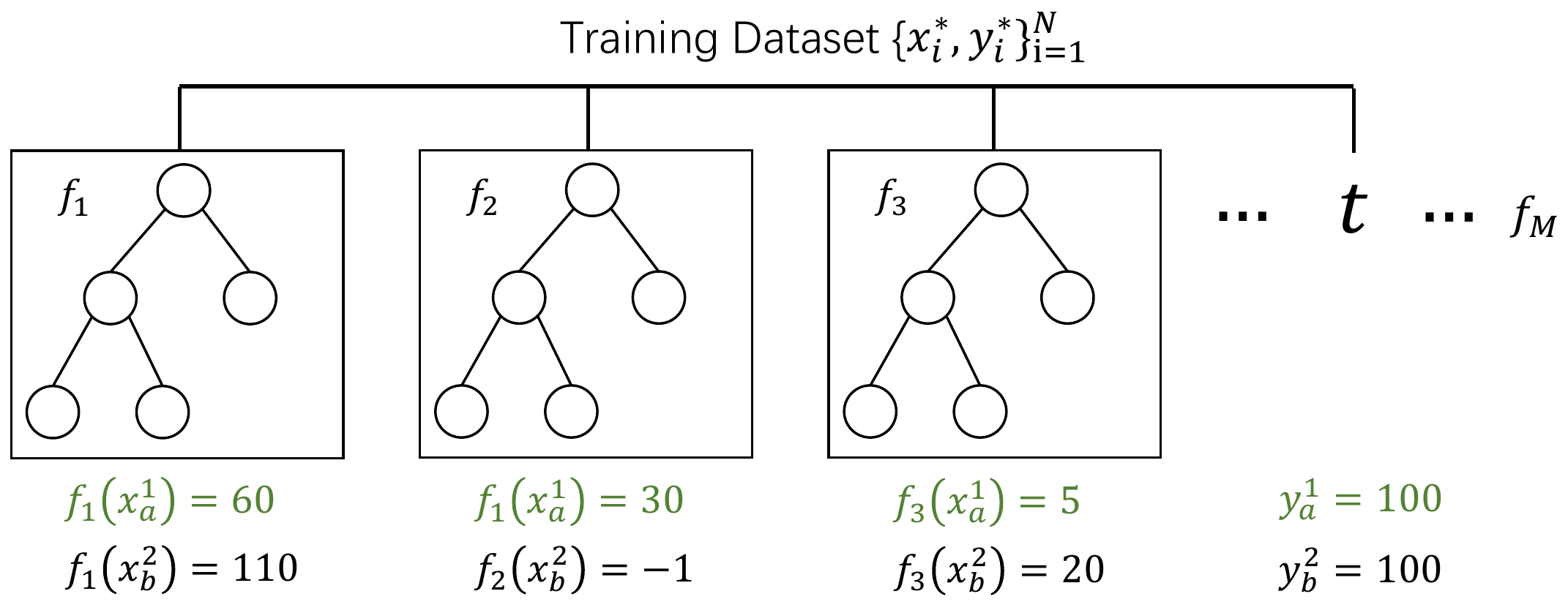}
\end{center}
  \caption{An example of the impact of local data heterogeneity on the performance of XGBoost model.}
  \label{Intuition}
\end{figure}

Consider the example illustrated in Fig. \ref{Intuition}. We have an XGBoost model consisting of $M$ trees in total where $f_t$ denotes the $t$-th tree, $t=1..M$. The XGBoost model is trained on the dataset $\{x_i^*, y_i^*\}_{i=1}^N$ for a regression task. We send this XGBoost model to two other clients and evaluate on their respective local datasets, $S_1$ and $S_2$. 

The prediction outcomes of the first three trees in the XGBoost tree ensemble on two data samples $\{x_a^1, y_a^1\} \in S_1$ and $\{x_b^2, y_b^2\} \in S_2$ are also labeled in Fig. \ref{Intuition}. Their ground truths are equal such that $y_a^1 = y_b^2 = 100$. Since local datasets $S_1$ and $S_2$ belong to two heterogeneous clients, the trees perform differently across data samples. For the first tree $f_1$, it gives a good initial prediction for $x_b^2$ (110) but not for $x_a^1$ (60). The second and third trees $f_2$ and $f_3$, on the contrary, sufficiently correct the residuals made by the first tree $f_1$ for $x_a^1$ (30, 5) but not for $x_b^2$ (-1, 20). In this case, a fixed learning rate (e.g., $\eta = 0.3$) may be too weak because ideally, we want a higher learning rate for $f_2(x_a^1)$ and $f_3(x_a^1)$ but a lower learning rate for  $f_2(x_b^2)$ and $f_3(x_b^2)$.

\vspace{1.0mm} \noindent \textbf{Moving towards the global optima} \hspace{1.0mm} As explained previously, data heterogeneity causes the trained XGBoost models on different clients' local datasets to converge to local optima that are far from each other. Consequently, given an unseen data sample, these XGBoost tree ensembles output different prediction results. However, among all XGBoost tree ensembles, some can give more accurate predictions because the unseen data sample may be closer to the underlying distribution of their trained datasets. Thus, applying a weighted sum on the diverse prediction results given by all XGBoost tree ensembles can lead to a more accurate final prediction value, helping us to move towards the global optima. 

It is important to point out that the approach of utilizing weighted sum to converge to the global optima is proved to be effective in the previous literature. FedAvg \cite{McMahanMRHA17} used the weighted sum of the aggregated model parameters according to the number of data samples presented in the clients' local datasets, and many kinds of literature have given theoretical convergence guarantees for the method \cite{KarimireddyKMRS20, li2019convergence}. Later FL strategies such as FedProx \cite{LiSZSTS20} also adopted the weighted sum of aggregated model parameters.

\subsection{Tree Ensembles Aggregation} \label{Section3.2}

\begin{figure}[t]
\setlength{\abovecaptionskip}{0.1cm}
\begin{center}
\includegraphics[width=0.95\linewidth]{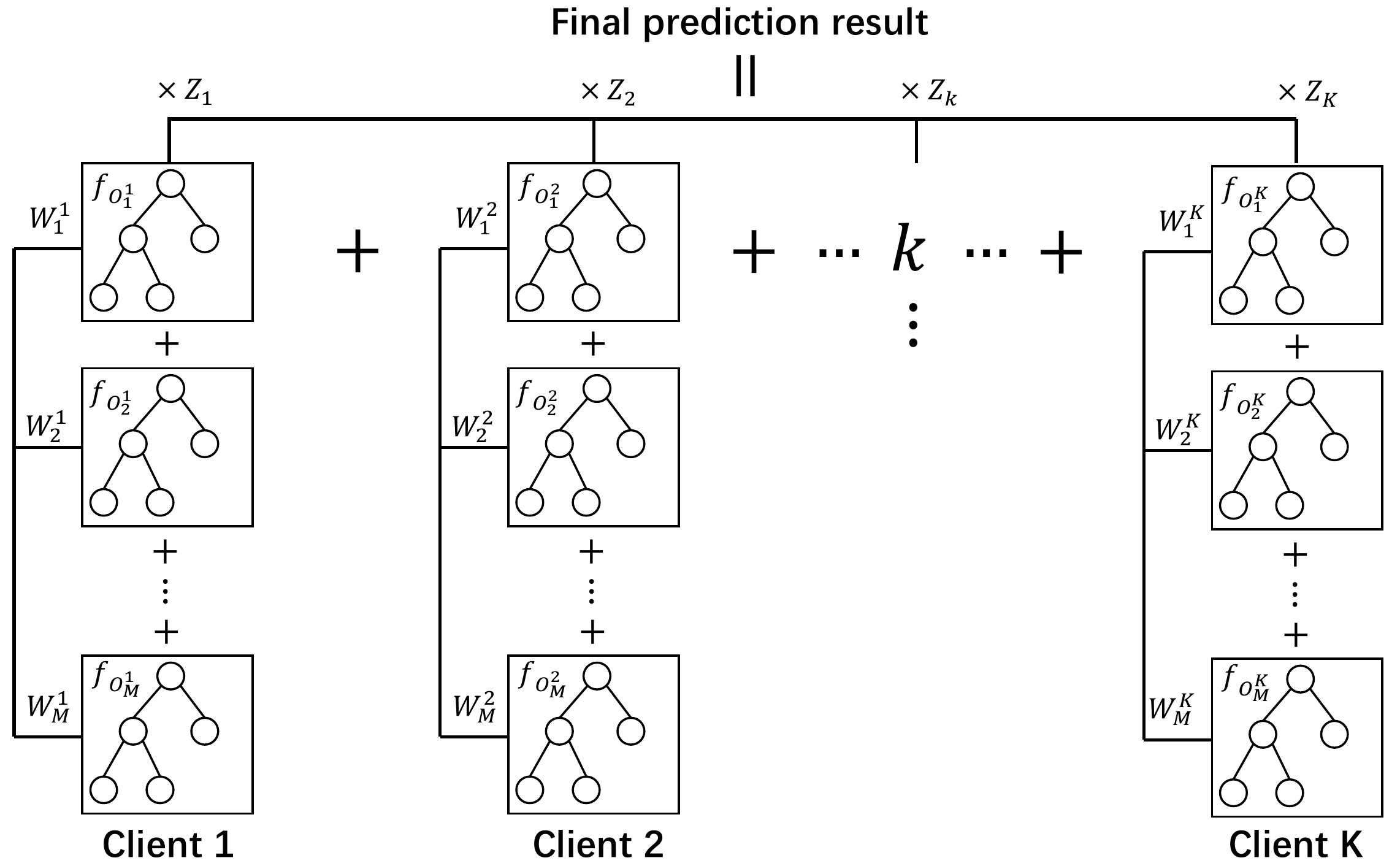}
\end{center}
  \caption{The aggregated tree ensemble. The final prediction given by the weighted sum of all trees.}
  \label{NN}
\end{figure}

Suppose there are $K$ clients participating in the training of federated XGBoost, and denote them as $(O_{1}, O_{2},...,O_{K})$. All clients' local datasets have different sample IDs but the same feature dimension $D$. Each client trains a XGBoost tree ensemble consisting of $M$ trees using its local dataset, where $f_{O_{t}^{k}}$ denotes the $t$-th tree constructed by client $k$, $t=1...M$ and $k=1...K$. To facilitate our intuitions, the final prediction result given an arbitrary data sample with feature dimension $D$ is calculated by the weighted sum of all trees from all clients as shown in Fig. \ref{NN}. Each vertical tree chain is the tree ensemble built by one client, where $W_{t}^{k}$ is the learning rate assigned to $f_{O_{t}^{k}}$ and $Z_k$ is the weight applied to the prediction result calculated by client $O_k$'s tree ensemble. We refer to this system as the aggregated tree ensemble. Both $f_{O_{t}^{k}}$ and $Z_k$ are learnable, which will be revealed in Section \ref{Section3.3}.

For all clients to calculate the final prediction result, each client needs to receive the aggregated tree ensemble with the help of the server. First, each client ensures that within its tree ensemble, all trees are sorted (i.e., if the tree ensemble is stored in an array, the $t$-th tree is at the $t$-th position). Then, as shown in Fig. \ref{Pipeline}(a), each client sends their built XGBoost tree ensemble and client ID ($CID = k$) to the server. The server sorts and concatenates all tree ensembles using $CID$s such that the $k$-th tree ensemble is always adjacent to both  $(k-1)$-th and $(k+1)$-th tree ensembles, as illustrated by the input layer of Fig. \ref{Pipeline}(b). Finally, the server broadcasts the sorted, aggregated tree ensembles to every client.

\subsection{Learnable Learning Rates by One-layer 1D CNN} \label{Section3.3}
\begin{figure}[t]
\setlength{\abovecaptionskip}{0.1cm}
\begin{center}
\includegraphics[width=1.0\linewidth]{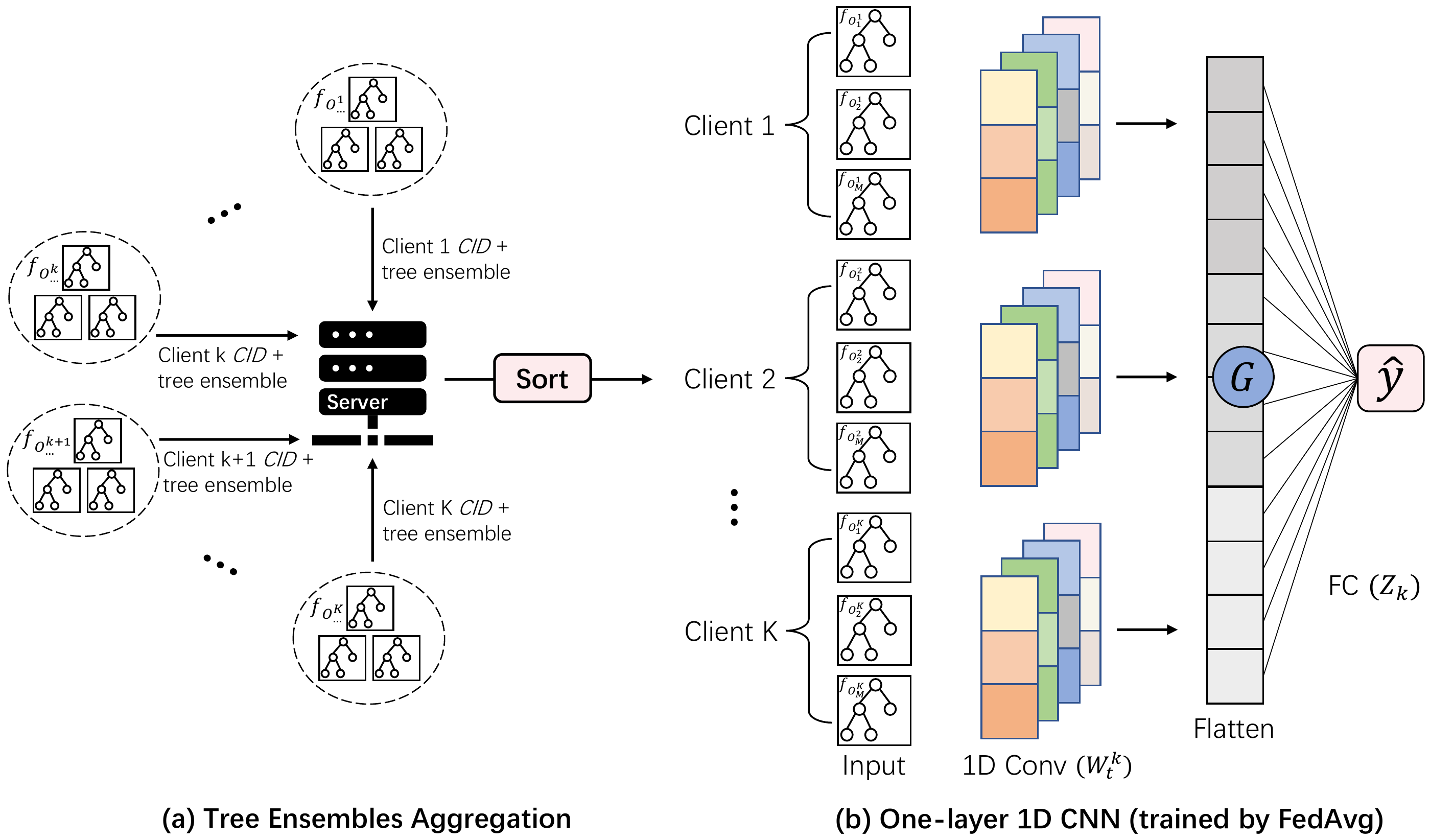}
\end{center}
  \caption{The pipeline. (a) tree ensembles aggregation and (b) one-layer 1D CNN to study the learning rates and output the final prediction result.}
  \label{Pipeline}
\end{figure}

We develop a method to learn the learning rate $W_{t}^{k}$ assigned to each tree $f_{O_{t}^{k}}$ by transforming the aggregated tree ensembles in Fig. \ref{NN} to a one-layer 1D CNN as shown in Fig. \ref{Pipeline}(b). In the first 1D convolution layer, the inputs are the prediction outcome of all trees. $G$ is the chosen activation function. 

\vspace{1.0mm} \noindent \textbf{Interpretability} \hspace{1.0mm} The small-sized model is interpretable. The kernel size and stride of the 1D convolution are equal to the number of trees, $M$, in each client's tree ensemble. Thus, each channel of the 1D convolution is the learnable learning rates ($W_{t}^{k}$) for all $f_{O_{t}^{k}}$ in the tree ensemble of a specific client $k$, and the number of convolution channels can be understood as the number of learning rate strategies that can be applied. The classification head, fully connected (FC) layer, contains the weighting factors ($Z_k$) to balance the prediction outcomes of each client's tree ensemble and calculate the final prediction result, which is also updated during training. The incentive for introducing activation $G$ is to avoid overfitting because a portion of the learned strategies will be deactivated. We set $G$ to be the most used activation function, ReLU.

\subsection{FedXGBllr} \label{Section3.4}

\begin{algorithm}[t]
\SetKwInput{Input}{Input}
\SetKwInput{Output}{Output}
\Input{A server $C$. K clients $(O_{1}, O_{2},...,O_{K})$, each with a local dataset $S_{k} \in \mathbb{R}^{N_{k} \times D}$. The number of trees to be trained $M$. FL training round $R$, local epoch $E$, local learning rate $\alpha$, local minibatch size $B$. Initial one-layer 1D CNN parameters $w_0$.}
\Output{Trained global federated XGBoost model}

\vspace{2.5mm}
\If{round $t = 0$ \label{round0_start}}{
    Each $O_k$ trains a XGBoost tree ensemble with $M$ trees using local dataset\;
    Each $O_k$ sends the trained XGBoost tree ensemble and $CID$ to $C$\;
    $C$ sorts and broadcasts the aggregated XGBoost tree ensemble ($K$ ensembles total, $MK$ trees total) with corresponded $CID$ to all $O_k$ ($K$ clients)\;
    $C$ initializes $w_0$ for the one-layer 1D CNN\;
    Each $O_k$ receives the aggregated XGBoost tree ensemble\;
    Each $O_k$ evaluates the aggregated XGBoost tree ensembles on local dataset $S_k \in \mathbb{R}^{N_{k} \times D}$ so that prediction outcomes of all $MK$ trees on all data samples, denoted as $S_k^* \in \mathbb{R}^{N_{k} \times MK}$, are acquired. \label{round0_finish}
}

\vspace{2.5mm}
\For{round $t = 1$ to $R$ \label{round1}}{
    $C$ selects all $K$ clients\;
    $C$ broadcasts $w_t$ to all $K$ clients\;
    \For{each client $O_k$ \textbf{in parallel}}{
        $w_{t + 1}^{k}$ = ClientUpdate($CID$, $w_t$, $E$, $B$, $\alpha$)\;
    }
    $w_{t+1} = \sum_{k=1}^{K} \frac{N_k}{N} w_{t+1}^k$\;
}

\vspace{2.5mm}
\SetKwFunction{ClientUpdate}{\textbf{ClientUpdate}}
\SetKwProg{Fn}{Def}{:}{}
\Fn{\ClientUpdate{$CID$, $w$, $E$, $B$, $\alpha$}}{
    $\beta$ = divide $S_{CID}^*$ into batches of size $B$\;
    \For{local epoch $e = 1$ to $E$}{
        \For{local batch $b \in \beta$}{
            $w = w - \alpha \Delta \ell(b; w)$\;
        }
    }
    \KwRet $w$\;
}
\caption{FedXGBllr.} 
\label{FedXGBllr}
\end{algorithm}
\setlength{\textfloatsep}{0.5em}
\setlength{\floatsep}{0.5em}

We introduce the new framework, \textbf{FedXGBllr}, to train a global federated XGBoost model by learning the learning rate for each tree with FL. The global federated XGBoost model consists of all clients' locally trained XGBoost tree ensembles and the globally trained one-layer 1D CNN. The detailed procedure is shown in Algorithm. \ref{FedXGBllr}. At round 0 (line \ref{round0_start} to \ref{round0_finish}), each client first trains its local XGBoost tree ensemble. The server then conducts tree ensemble aggregation and CNN initialization. After receiving the aggregated tree ensemble, all clients calculate the prediction outcomes given the aggregated tree ensemble on their local data samples. The calculated prediction outcomes are inputs of the CNN. It is worth noticing that the clients only build XGBoost models at round 0, and the aggregated tree ensemble is fixed after round 0. For the federated training of the one-layer 1D CNN after round 1 (line \ref{round1}), the protocol follows the standard FL algorithm, and we use FedAvg \cite{McMahanMRHA17}.

In FedXGBllr, the number of communication rounds is equal to the FL training rounds ($R$) because we send the trees (at round 0) and CNN's model parameters (after round 1). 

\section{Experiments}
In this section, we conduct extensive experiments to validate the effectiveness of our approach. We start by describing experiment setup and implementation details in Section \ref{Section 4.1}. We then discuss the experimental results with comparisons to the centralized baseline and the state-of-the-art method in Section \ref{Section 4.2}. Finally, we provide ablation studies and analysis to justify the interpretability and low communication overhead of our approach in Section \ref{Section 4.3}.

\subsection{Experiment Setup and Implementations} \label{Section 4.1}
\noindent \textbf{Comparison methods} \hspace{1.0mm}  We benchmark our method against one of the state-of-the-art and most influential works on horizontal federated XGBoost, SimFL~\cite{practical_LiWH20}, which adopts locality-sensitive hashing in the context of FL. Opposed to our method, SimFL trains the global XGBoost model by sharing the weighted gradients across rounds. We also use the centralized XGBoost trained on the whole dataset as baseline.

\vspace{1.0mm} \noindent \textbf{Dataset} \hspace{1.0mm}
Following SimFL~\cite{practical_LiWH20}, we evaluate our method on the same six tabular datasets for classification. We also conduct our experiment on four tabular datasets for regression. All datasets can be downloaded from LIBSVM data 
website~\footnote[1]{https://www.csie.ntu.edu.tw/~cjlin/libsvmtools/datasets/ \label{LIBSVM}}. 
The information of each dataset is summarized in Table. \ref{table:Datasets}. For all datasets, the training set to test set ratio is $0.75:0.25$ with random shuffling. The test set is used as the global test set at the server side. For the training set, we equally divide it according to the number of clients and assign the partitioned datasets to each client as their local dataset. 
SimFL~\cite{practical_LiWH20} only conducts experiments using 2 clients. We also provide the results of our method using 5 and 10 clients.

\begin{table}[t]
\setlength{\abovecaptionskip}{0.1cm}
\caption{Summary of datasets}
\begin{adjustbox}{width=1\linewidth,center} 
\setlength{\abovecaptionskip}{0.1cm}
  \footnotesize
  \centering
  \begin{tabular}{@{}llccc@{}}
  \toprule
  \textbf{Dataset} & \textbf{Task Type} & \textbf{Data No.} & \textbf{Dimension} & \textbf{Size}\\ 
  \hline \noalign{\vskip 1.0ex}

\textbf{a9a} & classification & 32,561 & 123 & 16MB
\\ \noalign{\vskip 1.0ex}

\textbf{cod-rna} & classification & 59,535 & 8 & 2.1MB
\\ \noalign{\vskip 1.0ex}

\textbf{ijcnn1} & classification & 49,990 & 22 & 4.4MB
\\ \noalign{\vskip 1.0ex}

\textbf{real-sim} & classification & 72,309 & 20,958 & 6.1GB
\\ \noalign{\vskip 1.0ex}

\textbf{HIGGS} & classification & 1,000,000 & 28 & 112MB
\\ \noalign{\vskip 1.0ex}

\textbf{SUSY} & classification & 1,000,000 & 18 & 72MB
\\ \midrule

\textbf{abalone} & regression & 4,177 & 8 & 253KB
\\ \noalign{\vskip 1.0ex}

\textbf{cpusmall} & regression & 8,192 & 12 & 684KB
\\ \noalign{\vskip 1.0ex}

\textbf{space\_ga} & regression & 3,167 & 6 & 553KB
\\ \noalign{\vskip 1.0ex}

\textbf{YearPredictionMSD} & regression & 515,345 & 90 & 615MB
\\
  \bottomrule
  \end{tabular}
  \label{table:Datasets}
\end{adjustbox}
\vspace{0.5em}
\end{table} 

\vspace{1.0mm} \noindent \textbf{Evaluation metric} \hspace{1.0mm} We report the performance on classification and regression datasets using Accuracy and Mean Squared Error (MSE) respectively, which is common practice.

\vspace{1.0mm} \noindent \textbf{Implementation details} \hspace{1.0mm} We use the Python package xgboost to train the local XGBoost models. Following SimFL~\cite{practical_LiWH20}, the maximum depth of all trees is set to 8. For our implementation, we set the number of trees in each tree ensemble to be 500 divided by the number of clients. The initial learning rate ($\eta$) is the same for all trees and is set to 0.1. Note that $\eta$ is a fixed hyperparameter of XGBoost (explained in Section~\ref{Section2}), and is not the learnable learning rates ($W_{t}^{k}$) that are refined by the one-layer 1D CNN (explained in Section~\ref{Section3.3}). For the globally trained one-layer 1D CNN and FL infrastructures, including clients and a server, we implement our method with PyTorch under Flower \cite{FlowerBeutel}, an end-to-end FL framework. For the CNN, we employ Kaiming initialization \cite{he2015delving} and set the number of convolution channels to 64. For each client, we train the CNN using Adam \cite{KingmaB14} with learning rate ($\alpha$) 0.001, $\beta_1$ momentum 0.5, and $\beta_2$ momentum 0.999. The local epoch ($E$) is set to 100, and the batch size ($B$) is set to 64. We train on one Nvidia Tesla V100 GPU.

\subsection{Experimental Results} \label{Section 4.2}

\begin{table}[t]
\setlength{\abovecaptionskip}{0.1cm}
  \footnotesize
  \caption{Quantitative results of FedXGBllr compared to SimFL and centralized baseline - \textbf{Accuracy $\uparrow$} (for the first six classification datasets), \textbf{MSE $\downarrow$} (for the last four regression datasets).}
  \begin{adjustbox}{width=1\linewidth,center}
  \setlength{\abovecaptionskip}{0.1cm}
  \footnotesize
  \centering
  \begin{tabular}{@{}lccccc@{}}
  \toprule
  \multirow{2}{*}{\textbf{Dataset}} & \multicolumn{3}{c}{\textbf{FedXGBllr}} & \textbf{SimFL~\cite{practical_LiWH20}} & \multirow{2}{*}{\specialcellC[2]{\textbf{Centralized} \\ \textbf{Baseline}}} \\ 
  
  \cmidrule(lr){2-4} \cmidrule(lr){5-5}
  
  & 2 clients & 5 clients & 10 clients & 2 clients & \\ \hline \noalign{\vskip 1.0ex}

\textbf{a9a} & 85.1 & 85.1 & 84.7 & 84.9 & 84.9 
\\ \noalign{\vskip 1.0ex}

\textbf{cod-rna} & 97.0 & 96.5 & 95.8 & 94.0 & 93.9 
\\ \noalign{\vskip 1.0ex}

\textbf{ijcnn1} & 96.3 & 96.0 & 95.3 & 96.4 & 96.3 
\\ \noalign{\vskip 1.0ex}

\textbf{real-sim} & 93.4 & 93.8 & 92.7 & 92.9 & 93.5
\\ \noalign{\vskip 1.0ex}

\textbf{HIGGS} & 71.5 & 70.9 & 70.3 & 70.7 & 70.7 
\\ \noalign{\vskip 1.0ex}

\textbf{SUSY} & 82.5 & 81.7 & 81.2 & 80.4 & 80.0
\\ \midrule

\textbf{abalone} & 3.6 & 4.4 & 4.9 & - & 1.3 
\\ \noalign{\vskip 1.0ex}

\textbf{cpusmall} & 8.0 & 8.5 & 9.5 & - & 6.7 
\\ \noalign{\vskip 1.0ex}

\textbf{space\_ga} & 0.024 & 0.033 & 0.034 & - & 0.024 
\\ \noalign{\vskip 1.0ex}

\textbf{YearPredictionMSD} & 80.3 & 82.7 & 91.6 & - & 80.5
\\

  \bottomrule
  \end{tabular}
  \label{table:Results}
\end{adjustbox}
\vspace{0.5em}
\end{table} 

Table. \ref{table:Results} demonstrates the quantitative results of FedXGBllr. The number of communication rounds ($R$) is set to 10. For all experiments, we take an average of 5 runs. From the results, our approach outperforms or reaches comparable accuracy to SimFL~\cite{practical_LiWH20} and the centralized baselines on all six classification datasets with 2 clients. For the regression datasets, our approach achieves comparable or slightly higher MSE compared to the centralized baseline.  

For both classification and regression datasets, our method performs better on larger datasets. We hypothesize this is due to the generalization capability of CNN scaling up with the volume of data. Additionally, as the number of clients increases from 2 to 5 and 10, the performance slightly decreases. We think it is reasonable because FL is harder with more clients \cite{Kairouz21_Open_Problems}. 

The results suggest 10 rounds are sufficient for FedXGBllr to build a good global federated XGBoost model. However, it is worth mentioning that the number of rounds needed to reach good performance correlates with the number of local epochs ($E$) to train the one-layer 1D CNN on the client side. A higher $E$ may require fewer communication rounds (consider the extreme when $E=1$). Our implementation uses $E=100$, and leaves the optimal trade-off of $E$ and $R$ for future studies.

\subsection{Ablation Studies and Analysis} \label{Section 4.3}

\noindent \textbf{Communication overhead} \hspace{1.0mm} We compare the total communication overhead to build a global federated XGBoost model of our approach to baseline, SimFL~\cite{practical_LiWH20}. For all comparisons, we assume the number of clients $K$ to be 10 and the total number of built XGBoost trees $M$ to be 500 with a depth $L$ of 8 in order to be consistent with SimFL's efficiency experiments. The communications overhead of our approach is independent of the dataset size $N$, and can be expressed as: 
\begin{equation}\label{eq6}
2K(M \times SZ\_t + R \times SZ\_nn)
\end{equation}
\noindent where $R$ is the number of FL training rounds, $SZ\_t$ is the size of each tree in bytes, and $SZ\_nn$ is the size of the one-layer 1D CNN in bytes. Therefore, $2KM \times SZ\_t$ is the communication overhead during tree ensembles aggregation at round 0, and $2KR \times SZ\_t$ is the communication overhead of the federated training of the CNN from round 1 to $R$. We assume $R$ to be 10 because this number is sufficient for our approach to reach good performance as explained in Section \ref{Section 4.2}. $SZ\_nn$ is 0.03MB (Table. \ref{table:size}). In practice, $SZ\_t$ is negligible as the size of 500 trees built by the xgboost package is only 48 bytes.

\begin{wraptable}{R}{3.2cm}
\setlength{\abovecaptionskip}{0.1cm}
\vspace{-3.5mm}
\footnotesize
\caption{Comparison of communication overhead (MB).}
\begin{adjustbox}{width=3.6cm, right} 
\setlength{\abovecaptionskip}{0.1cm}
  \centering
  \begin{tabular}{@{}lcc@{}}
  \toprule
   \textbf{Dataset} & \textbf{FedXGBllr} & \textbf{SimFL} \cite{practical_LiWH20} \\ 
  \hline \noalign{\vskip 1.0ex}

\textbf{a9a} & 6.0 & 150.4 ($25 \times$)
\\ \noalign{\vskip 1.0ex}

\textbf{cod-rna} & 6.0  & 249.3 ($42 \times$)
\\ \noalign{\vskip 1.0ex}

\textbf{ijcnn1} & 6.0  & 218.4 ($36 \times$)
\\ \noalign{\vskip 1.0ex}

\textbf{real-sim} & 6.0  & 323.1 ($54 \times$)
\\ \noalign{\vskip 1.0ex}

\textbf{HIGGS} & 6.0  & 4216 ($703 \times$)
\\ \noalign{\vskip 1.0ex}

\textbf{SUSY} & 6.0  & 4136 ($689 \times$)
\\
  \bottomrule
  \end{tabular}
  \label{table:overhead}
\end{adjustbox}
\end{wraptable} 

The communication overhead of SimFL~\cite{practical_LiWH20} in bytes is given by $8KN \times Hash + 8M[N + (2^{L}-1)(K-1)]$, where $Hash$ is the number of hash functions. Table. \ref{table:overhead} illustrates the comparison of the total communication overhead. We include the results using the six classification datasets because SimFL~\cite{practical_LiWH20} provided the exact values on them. We can see the communication overhead of our approach is significantly lower especially as the dataset size scales up. We save the communication cost by at least a factor of 25, and can save up to a factor of 700.

Although we did not compare the exact numbers, our communication overhead is also significantly lower than encryption-based methods such as FedXGB \cite{LiuMLMND020}, whose training shares encryption keys and communication cost scales up linearly with both input size and the number of clients. 

\vspace{1.0mm} \noindent \textbf{Model interpretability}
\hspace{1.0mm} We want to know if the interpretability of our one-layer 1D CNN couples with the high performance? We change the first 1D convolution layer with kernel size and stride equal to the number of trees in each client tree ensemble with: 1) standard convolution with kernel size 3, stride 1, and 2) FC layer with dimension 256, and remove the flattened layer. The number of communication rounds is set to 10. We fix the number of clients to be 5. The results are shown in Table. \ref{table:models}. We also show the number of parameters and total size of each model in Table. \ref{table:size}.

From the results, our one-layer 1D CNN reaches the best performance on all datasets although it has the smallest number of parameters and total size. This suggests the effectiveness and interpretability of our CNN model. We find that for all datasets, the performance gap between our interpretable CNN and 2-layer FCNN is much larger than the gap between our interpretable CNN and CNN with standard kernel size and stride. Also, the gap exaggerates as the dataset size increases. We argue that in addition to our reasoning in Section \ref{Section3.3}, it is because CNN can leverage the temporal information across the tree ensembles built by the clients, and our interpretable CNN has the right amount of temporal resolution (i.e., with kernel size = stride). 

\begin{table}[t]
\setlength{\abovecaptionskip}{0.1cm}
\caption{Model interpretability. $k = kernel\_size$, $s = stride$, $n = client\_tree\_num = 500 / client\_num$.}
\begin{adjustbox}{width=1\linewidth,center} 
\setlength{\abovecaptionskip}{0.1cm}
  \footnotesize
  \centering
  \begin{tabular}{@{}lccc@{}}
  \toprule
  \textbf{Dataset} & \specialcellC[2]{\textbf{1-layer 1D CNN} \\ \textbf{(k = s = n)}} & \specialcellC[2]{\textbf{1-layer 1D CNN} \\ \textbf{(k = 3, s = 1)}} & \textbf{2-layer FCNN}\\ 
  \hline \noalign{\vskip 1.0ex}

\textbf{a9a} & 85.1 & 83.9 & 82.8
\\ \noalign{\vskip 1.0ex}

\textbf{cod-rna} & 96.5 & 96.3 & 94.7
\\ \noalign{\vskip 1.0ex}

\textbf{ijcnn1} & 96.0 & 95.1 & 92.2
\\ \noalign{\vskip 1.0ex}

\textbf{real-sim} & 93.8 & 93.2 & 91.6
\\ \noalign{\vskip 1.0ex}

\textbf{HIGGS}  & 70.9 & 70.5 & 67.9
\\ \noalign{\vskip 1.0ex}

\textbf{SUSY}  & 81.7 & 81.3 & 77.5
\\ \midrule

\textbf{abalone} & 4.4 & 4.4 & 5.8
\\ \noalign{\vskip 1.0ex}

\textbf{cpusmall}  & 8.5 & 9.2 & 12.6
\\ \noalign{\vskip 1.0ex}

\textbf{space\_ga}  & 0.033 & 0.034 & 0.044
\\ \noalign{\vskip 1.0ex}

\textbf{YearPredictionMSD} & 82.7 & 87.5 & 117.7
\\
  \bottomrule
  \end{tabular}
  \label{table:models}
\end{adjustbox}
\vspace{0.5em}
\end{table} 

\begin{table}[t]
\setlength{\abovecaptionskip}{0.1cm}
\caption{Number of parameters and size of the models. $k = kernel\_size$, $s = stride$, $n = client\_tree\_num = 500 / client\_num$.}
\begin{adjustbox}{width=1\linewidth,center} 
\setlength{\abovecaptionskip}{0.1cm}
  \footnotesize
  \centering
  \begin{tabular}{@{}lccccc@{}}
  \toprule
   & \multicolumn{3}{c}{\textbf{1-layer 1D CNN (k = s = n)}} & \multirow{2}{*}{\specialcellC[2]{\textbf{1-layer 1D CNN} \\ \textbf{(k = 3, s = 1)}}}  & \multirow{2}{*}{\textbf{2-layer FCNN}}\\
   \cmidrule(lr){2-4}
   & 2 clients & 5 clients & 10 clients \\
  \hline \noalign{\vskip 1.0ex}

\textbf{Total params} & 16193 & 6,785 & 3905 & 32,257 & 128,513
\\ \noalign{\vskip 1.0ex}

\textbf{F/b pass size (MB)} & 0.01 & 0.01 & 0.01 & 0.73 & 0.01
\\ \noalign{\vskip 1.0ex}

\textbf{Params size (MB)} & 0.06 & 0.03 & 0.02 & 0.12 & 0.49
\\ \noalign{\vskip 1.0ex}

\textbf{Total size (MB)} & 0.07 & 0.04 & 0.03 & 0.86 & 0.50
\\
  \bottomrule
  \end{tabular}
  \label{table:size}
\end{adjustbox}
\vspace{0.5em}
\end{table} 

\section{Conclusion and Future Works}
We propose a novel framework, FedXGBllr, for horizontal federated XGBoost which does not rely on the sharing of gradients and hessians. Extensive evaluations prove our approach is robust, interpretable, and communication efficient. Specifically, we reach performance comparable to state-of-the-art method and reduce communication cost with factors ranging from 25x to 700x. We use FedAvg \cite{McMahanMRHA17} in this work to learn the learnable learning rates by training a small one-layer 1D CNN. It is important to point out that more advanced FL training algorithms can also be applied and better performance may be achieved; however, we leave it for future studies as it is not the focus of this research. Future works also include extending FedXGBllr to vertical setting. 



\newpage
\bibliographystyle{ACM-Reference-Format}
\bibliography{egbib}










\end{document}